\documentclass[letterpaper, 10 pt, conference]{ieeeconf}  % Comment this line out if you need a4paper

\IEEEoverridecommandlockouts                              % This command is only needed if 
\usepackage{graphics} % for pdf, bitmapped graphics files
\usepackage{epsfig} % for postscript graphics files
\usepackage{mathptmx} % assumes new font selection scheme installed
\usepackage{times} % assumes new font selection scheme installed
\usepackage{amsmath} % assumes amsmath package installed
\usepackage{amssymb}  % assumes amsmath package installed
\usepackage{cite}
\usepackage{url}
\usepackage[hidelinks]{hyperref} % Not working for... reasons?
\usepackage{tikz}
\usepackage{nicematrix}
\usepackage[table]{xcolor}
\usepackage{graphicx} % Required for \resizebox

\newcommand{\eg}{\textit{e}.\textit{g}.}
\newcommand{\etc}{\textit{etc}.}

\title{\LARGE \bf
REBAR: Reference Ethical Benchmark for Autonomy Readiness
}

\author{Jonathan Diller$^{1}$, David Barnes$^{*2}$, Rebekah Bogdanoff$^{*4}$, Rhett Collier$^{*4}$, Roddy Collins$^{*3}$, Keith Fieldhouse$^{*3}$,\\%
Yonatan Gefen$^{*3}$, Cameron Johnson$^{*3}$, Anuriha Kodali$^{*1}$, Brad Kriel$^{*4}$, Varun Murali$^{*5}$,  James Niehaus$^{*6}$,\\%
Mish Sukharev$^{*4}$, Joseph VanPelt$^{*3}$, Anthony Hoogs$^{3}$, Vijay Kumar$^{1}$, Arslan Basharat$^{3}$%
\thanks{\hspace*{-\parindent}$^{*}$ listed in alphabetical order.\newline
$^{1}$ University of Pennsylvania; $^{2}$ David Barnes, LLC; $^{3}$ Kitware, Inc.\newline 
$^{4}$ Duality Robotics, Inc.; $^{5}$ Texas A\&M University\newline 
$^{6}$ Charles River Analytics\newline 
Correspondence: {\tt\small diller@seas.upenn.edu}, {\tt\small arslan.basharat@kitware.com}\newline%
Approved for public release: distribution is unlimited.}%
}

\begin{document}

\maketitle
\thispagestyle{empty}
\pagestyle{empty}

%%%%%%%%%%%%%%%%%%%%%%%%%%%%%%%%%%%%%%%%%%%%%%%%%%%%%%%%%%%%%%%%%%%%%%%%%%%%%%%%
\begin{abstract}
As autonomous systems grow more advanced, objective metrics to evaluate their ethical and legal compliance are critical for informing end users of their limitations and ensuring accountability of those who misuse them. Current ethical embodied AI frameworks remain mostly qualitative, focusing on system design (through safety guardrails or targeted red teaming), and the realized guardrails often directly disallow unsafe behavior without providing the user with an override or interpretable reason. Instead, there is a need for computable metrics through rigorous testing that allow a user to determine the applicability of the system to the task. To address this gap, we introduce the Reference Ethical Benchmark for Autonomy Readiness (REBAR), a quantitative test and evaluation framework for autonomous systems. REBAR maps operating metrics into a computable Autonomy Readiness Level (ARL) rubric that can quantify ethical performance. Key innovations of the framework include a neuro-symbolic Large Language Model (LLM) approach to calculate and explain the ethical difficulty of scenarios, LLM-driven at-scale generation of test instances, and a versatile, photorealistic simulation environment. By evaluating white-box autonomy solutions through this rigorous testing pipeline, REBAR delivers an objective and repeatable benchmark score, bridging the gap between abstract principles and verifiable, accountable autonomy.
\end{abstract}

%%%%%%%%%%%%%%%%%%%%%%%%%%%%%%%%%%%%%%%%%%%%%%%%%%%%%%%%%%%%%%%%%%%%%%%%%%%%%%%%
\section{Introduction}
Safety in the era of rapidly evolving foundation models (FMs), agentic AI, and embodied AI is a critical challenge. 
% The path to translating general tasks into execution on embodied systems is narrowing, but it has brought challenges in the representational power of FMs and in regulating hallucinations and unsafe behavior in agents, which can have broader impacts when applied to autonomous systems.
% Current research focuses on mitigating harm in a general setting and building guardrails for systems \cite{gdm2024autort,sermanet2025asimov1,sermanet2025scifi}.
% However, guardrails that are ethically relevant tend to be subjective compared to those that are legally relevant and the execution of such guardrails become dependent on complex environmental context.
Current research has explored safety and evaluation for autonomous and AI systems through guardrails, red-teaming, and benchmark-driven frameworks\cite{gdm2024autort,sermanet2025asimov1,sermanet2025scifi,HELM2022,GAIA2023}, as well as approaches for embedding normative or procedural reasoning in embodied agents \cite{PARTNR, Bai2022ConstitutionalAI, Ahn2022SayCan}. 
These efforts have advanced robustness, alignment, and decision-making capabilities, but largely focus on functional performance, policy enforcement, dataset creation, or internal reasoning. Ethical AI (EAI) assessment frameworks, such as \cite{Hendrycks2021ETHICS,DIU2025,CDAO2025,Tabassi2023}, are predominantly qualitative, do not provide measurable, objective, computable metrics, and their primary focus is often on system design/dataset creation rather than testing. 
Furthermore, the decomposition of ethical principles into quantified measurements can be subjective, ambiguous, and variable based on specific use cases or scenarios. 
As autonomous systems become increasingly context-adaptive and operationally impactful, humankind requires principled, repeatable approaches to assess whether these systems behave in ways consistent with ethical, legal, and mission-aligned requirements as expressed and implied by the operator.
Beyond guiding operator decisions, the deployment of autonomous systems increasingly demands accountability mechanisms grounded in quantitative performance metrics.
The proposed work builds upon foundational work in ethical DevOps \cite{Taddeo2023,Alessandro2025Artificial} and ethical AI metrics \cite{Hu2022Doppelganger,Hu2024,Hoffman2023,Klein2015}. 
% However, ethically motivated guardrails and reasoning frameworks are often specified through subjective policies or static rules, and do not provide a computable, scenario dependent measure of ethical behavior. Existing benchmarks similarly emphasize task success or alignment proxies without capturing context dependent ethical tradeoffs.

% This motivates the need for a framework that evaluates ethical behavior directly, as a function of scenario context and observable system outcomes. To address these challenges, we propose the Reference Ethical Benchmark for Autonomy Readiness (REBAR), a quantitative benchmark to measure the ethical performance of an autonomous system. The framework's objective is to challenge the system with carefully conceived scenarios that thoroughly cover application relevant behavior through an ethical lens. The result is a quantified, relevant, comprehensive, and repeatable benchmark score of ethical behavior in the context of the user's application. REBAR maps measurable performance to ethical readiness levels, enabling repeatable, quantitative, and context sensitive assessment in the context of tasks and environments that are relevant to that user.

This motivates the need for a framework that evaluates ethical behavior directly, as a function of scenario context and observable system outcomes. To address these challenges, we propose the Reference Ethical Benchmark for Autonomy Readiness (REBAR). 
% REBAR challenges autonomous systems with carefully conceived scenarios that strain application-relevant behavior through an ethical lens. 
REBAR strains an autonomous system's application-relevant behavior through carefully conceived scenarios and quantifies performance through an ethical lens. 
% By mapping measurable performance to ethical readiness levels, the framework provides a comprehensive, repeatable, and quantitative benchmark score tailored to the user's specific operational environment.

% This motivates the need to evaluate ethical behavior directly through scenario context and observable outcomes. To address this gap, we propose the Reference Ethical Benchmark for Autonomy Readiness (REBAR). REBAR assesses autonomous systems using carefully conceived, application-relevant scenarios to map measurable performance to ethical readiness levels. The result is a quantitative, repeatable benchmark score that provides a context-sensitive assessment of ethical behavior.

% In this work, we employ the idea of testing and evaluation of systems to inform a given user about the ``readiness" of the system in the context of tasks and environments that are relevant to that user.

\begin{figure*}[!ht]
    \centering
    \includegraphics[width=0.9\textwidth]{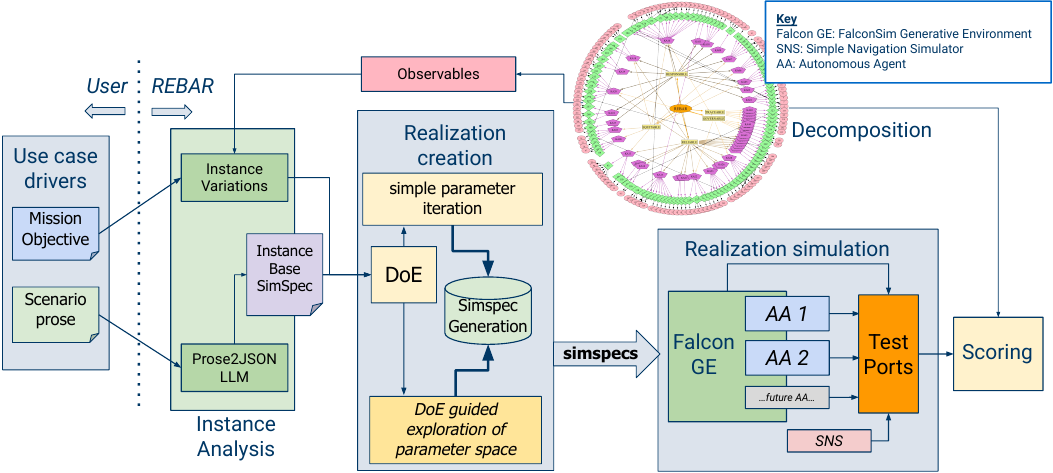}
    \caption{Proposed REBAR framework. A user can simply define a mission objective along with a natural language description of a scenario. Given this user-provided input and the relevant observables from the ethical decomposition graph, REBAR's instance analysis and realization creation modules can produce thousands of instance variations of the user's scenario as SimSpecs. These SimSpecs configure unique variants of the Falcon GE (or simulation environment), with a range of predicted ethical difficulties based on the combinations of parameters. The autonomy is scored based on its adherence to the decomposition graph's mapping of the ethical framework to the observables relevant to the scenario.}
    \label{fig:rebar}
\end{figure*}

The REBAR framework is designed to be modular and adaptable to different use cases of automated benchmarking, testing, and evaluation of AI and autonomy. 
Inspired by the Values, Criteria, Indicators, and Observables (VCIO) framework~\cite{Fetic2020Principles}, REBAR provides a novel, graphical decomposition of the mission under test into computable evaluation tasks, resulting in consolidated performance metrics reported as Autonomy Readiness Levels (ARL).
The REBAR pipeline (see Fig.~\ref{fig:rebar}) executes computable tests at scale by utilizing large language models (LLMs) for natural language processing, design-of-experiments over the simulation parameters, 
automated rendering in the simulation environment, and online autonomous agents whose outputs are scored across thousands of test realizations. Finally, the benchmark ARL scores are consolidated over the decomposition graph, characterizing the performance of an autonomous agent.  
The key innovations of the REBAR approach include:
\begin{itemize}
    % \item A semi-automated capability to transform mission parameters, user intent, and ethical values into a graphical, computable representation for generating computable tests at scale.
    \item A semi-automated capability to transform mission parameters, user intent, and ethical values into a graphical and quantitative representation for at-scale testing.
	\item An ARL score computation approach to measure the operational and ethical competence of autonomous agents that models the ethical difficulty of the scenario and decomposes performance into observables, values-actions-behaviors, key attributes, and principles.
    \item Online integration of the simulation environment \cite{Falcon2025} with multiple agents, instance generation stage, and the scoring stage via test ports to conduct automated tests at scale.
\end{itemize}

\section{Ethical Autonomous Readiness Levels}
\label{sec:ARLs}
To transition from abstract ethical ideals to computable evaluation metrics, we need a set of foundational criteria to evaluate, quantify, and compare the ethical maturity of varying autonomous capabilities.
We make use of the criterion of \cite{board2019ai} as the basis for determining the ARLs.
%Adapting from the foundational AI Ethical Principles defined by the U.S. Department of Defense, we define a set of ARLs as dimensional axes of qualitative performance. 
These bases are:
\begin{itemize}
	\item \textbf{Responsible:} Users shall exercise appropriate levels of judgment and care, while remaining responsible for the development, deployment, and use of AI capabilities. 
	\item \textbf{Equitable:} Organizations must take deliberate steps to minimize unintended bias in AI capabilities. 
	\item \textbf{Traceable:} An organization's AI capabilities must be developed and deployed such that relevant personnel possess an appropriate understanding of technology, development processes, and operational methods applicable to AI capabilities, including transparent and auditable methodologies, data sources, and design procedure and documentation. 
	\item \textbf{Reliable:} An organization's AI capabilities must have explicit, well-defined uses, and the safety, security, and effectiveness of such capabilities will be subject to testing and assurance within those defined uses across AI capabilities' entire life-cycle. 
	\item \textbf{Governable:} Organizations must design and engineer AI capabilities to fulfill their intended functions while possessing the ability to detect and avoid unintended consequences, and the ability to disengage or deactivate deployed systems that demonstrate unintended behavior.
\end{itemize}

To ensure that autonomous systems can be objectively evaluated against these standards, these principles must be conceptualized not as binary compliance checkboxes, but as continuous metrics of readiness. There is an immediate need for a comprehensive testing framework capable of quantitatively scoring an autonomous agent within each of these specific categories to achieve operational accountability and trust. Such a framework must translate these high-level ARL dimensions into observable indicators, challenge the systems with complex ethical dilemmas, and generate computable scores during simulated, high-stakes scenarios.
% Treating these tenets as bases allows researchers, engineers, and end users to establish a comparative baseline. 
% This paradigm shift enables the direct comparison of the ethical behavior, architectural safeguards, and mission alignment of different autonomous agents.

\begin{figure*}[!ht]
    \centering
    \includegraphics[width=0.9\textwidth]{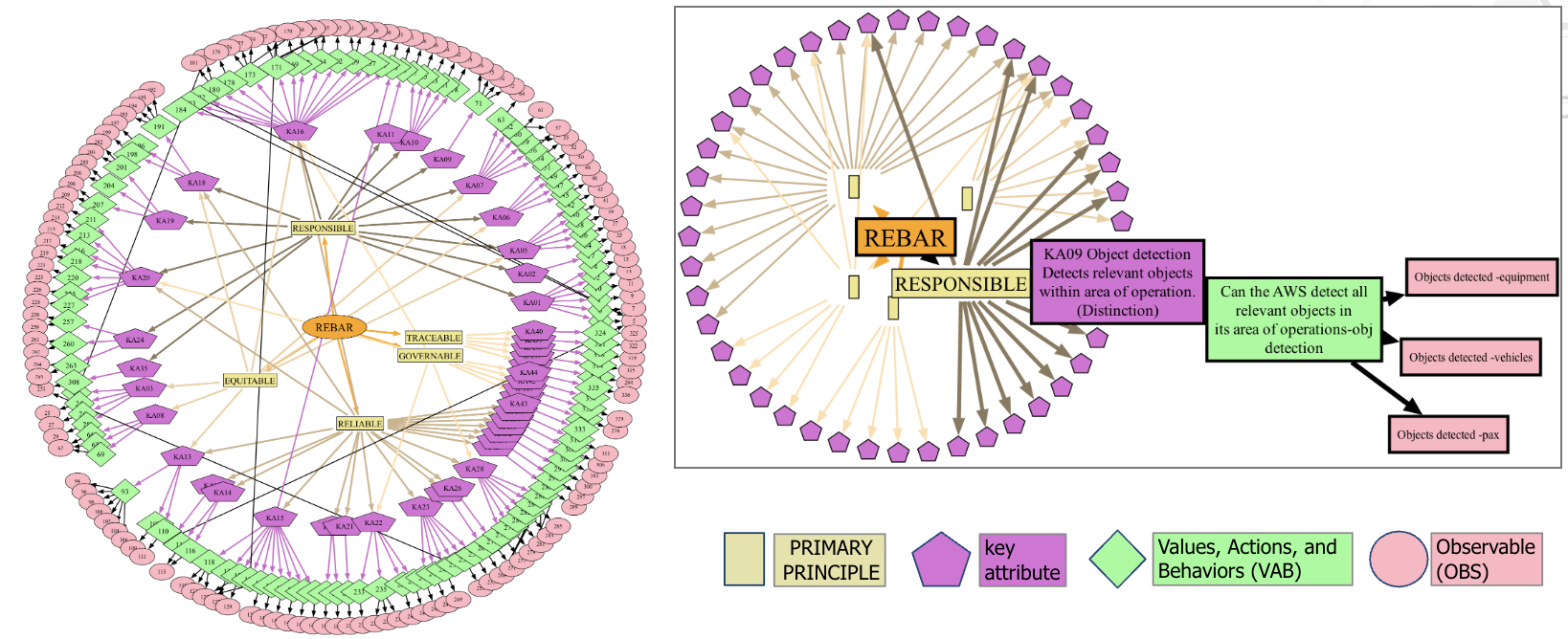}
    \caption{REBAR graph visualization mapping principles, key attributes, VABs, and observables. The ethical decomposition grounds each principle to a combination of observables intended to be quantifiable, scenario-relevant, and comprehensive.}
    \label{fig:rebar_graph}
\end{figure*}

\section{Proposed Benchmark: The REBAR Framework}
To quantify how a given autonomous system performs across each ARL base, we propose REBAR. 
This pipeline decomposes the bases into measurable observables. These observables, along with user mission objectives and scenario descriptions, go through an Instance Analysis stage where they are compiled into a self-contained base simulation specification (SimSpec), detailing the simulation environment, the autonomous system's capabilities, and the spatial constraints of all actors. 
A Realization creation stage takes the base SimSpec and composes several variations of the simulation setup using a Design of Experiments (DoE) methodology. The result is a set of SimSpecs that each specify a unique simulation configuration. The framework runs each SimSpec in a Realization Simulation stage, where the autonomous agent under test runs the given mission. The measurable observables are polled from each simulation run through test ports and composed into the ARL scores. Each of these stages is discussed further in the following subsections.
% The REBAR pipeline operates through a comprehensive, structured workflow: hierarchical value decomposition, automated scenario generation, executable simulation instantiation, and iterative statistical evaluation driven by a Design of Experiments (DOE) methodology.

\subsection{Hierarchical Ethical Decomposition}
\label{sec:decomposition}
REBAR translates abstract qualitative guidelines into measurable performance indicators using a four-level structured decomposition process inspired by the VCIO framework~\cite{Fetic2020Principles}. These four levels are:

\begin{enumerate}
    % \item Principles:  Responsible AI Principles (RAIPs) 
    \item Principles: Responsible AI Principles (RAIPs) listed above.
    \item Key Attributes (KAs): Testable criteria that define adherence to each Principle within a mission context (e.g., object detection).
    \item Values-Actions-Behaviors (VABs): System behaviors that represent legal and ethical alignment (e.g., detecting task-relevant objects).
    \item Observables (OBS): The quantified telemetry and simulation data extracted during testing (e.g., detecting a specific vehicle).
\end{enumerate}

The decomposition takes the form of a Directed Acyclic Graph (DAG), where the REBAR node acts as the source to the five RAIPs. Each principle node is the root of multiple KAs, and so on, down to Observable leaves. Figure~\ref{fig:rebar_graph} shows an example of this decomposition. The observables directly populate the ARL rubric; their use in computing the ARL score is described in Section~\ref{sec:arl_scoring} below. The ARLs can be computed at the Principles level as well as lower levels like KAs, as shown in Fig.~\ref{fig:arl_results}.

\subsection{Instance Analysis}
The Instance Analysis stage uses mission objectives, written scenario descriptions, and the decomposed observables to generate base SimSpecs. REBAR employs an LLM-driven package, Prose2JSON, that converts natural-language descriptions directly into these base specifications. Prose2JSON inherently understands parameterized text (e.g., creating distinct variations for ``[1,2] vehicles [20-40] meters away'').

Each base SimSpecs represents a scenario with a unique ethical difficulty level and enables the framework to generate massive sets of ``equi-ethical'' scenario variations. The base SimSpecs are handed to the Realization Creation stage to be used as a basis for generating scenario variations for the given mission and scenario.

\subsection{Realization Creation}\label{sec:3b}
% Given the vast parameter space created by environmental and ethical variables, the Realization Creation stage implements a rigorous DoE methodology to maximize the information gathered while optimizing computational resources. Rather than executing purely exhaustive combinations, the DoE pipeline utilizes an iterative sampling strategy. Initial screening relies on space-filling designs, such as Nearly Orthogonal Latin Hypercubes (NOLH), to ensure uniform coverage with minimal factor correlation. 
The Realization Creation stage takes the base SimSpec and generates a large set of SimSpec realizations for the given scenario. This is handled by the ``REBAR-Orchestrator'', a structured package that allows test engineers precise control over highly dimensional, interdependent parameter sets, expanding baseline experiments into tens of thousands of unique realizations. 
Falcon GE produces a high-fidelity simulation for online testing of autonomy for a given scenario, while SNS is designed to generate a low-fidelity 2D overhead simulation of that scenario, along with a surrogate model that replaces the autonomy.
Similarly, a simulated computer vision module can optionally replace the autonomy's real pixel-based object detection and tracking with perfect or programmatically degraded ground truth, to ablate vision issues from reasoning performance.

To induce varying levels of ethical difficulty, REBAR manipulates various parameters such as weather, GPS degradation, or bystander density. The framework uses a system of ``ethical tensions'' to describe an increasing level of difficulty for the autonomous system to accomplish a task, on a scale of 1 to 5. For example, a UAV locating a task-relevant object in an open field on a clear day would be a Tension 1-type scenario, while identifying the object in a cluttered environment on a rainy day would be a Tension 5 scenario.

\subsection{Realization of Simulation}\label{sec:3c}
In this phase, each generated SimSpec is loaded into the simulation. 
Each simulation starts with a base environment.
A constraint solver then ``places" the objects that need to be varied within pre-specified relational constraints. 
The solver validates that no ``placed" objects intersect, and when the constraints are satisfied, the simulation proceeds. 
At fixed periodic intervals during the simulation, the autonomous agent and the simulation engine produce logs (\eg, telemetry, detections \etc) which are used for scoring.

\subsection{ARL Scoring}
\label{sec:arl_scoring}
The REBAR framework calculates an ARL score and confidence ratio for each simulation run. These scores are calculated from the Observable leaves up to the main Principals of the ethical decomposition DAG (introduced in Section~\ref{sec:decomposition}). The framework employs the Principle of Minimal Ethical Difficulty: \textit{An Observable can only certify an agent to the \underline{minimum} of its configured ethical difficulties}.

\begin{table}[t]
\centering
\caption{Key Factor Table Example for Vehicle Detection}
\label{tab:key_factors}
\scriptsize % Stepping down one more size to fit the single column
\setlength{\tabcolsep}{2.5pt} % Reduces the horizontal whitespace between columns
\begin{NiceTabular}{p{2.2cm} rrrrrrrrrr}[hvlines, rules/color=gray!30, cell-space-limits=3pt]
\textbf{Observable} & \multicolumn{10}{c}{Object Detection: classifying vehicles} \\
\textbf{Key Factors} & 10 & 20 & 30 & 40 & 50 & 60 & 70 & 80 & 90 & 100 \\
Hours till solar noon & 0 & 0.89 & 1.78 & 2.67 & 3.56 & 4.44 & 5.33 & 6.22 & 7.11 & 8 \\
Rain level & 0 & 1 & 2 & 4 & 4 & 5 & 6 & 7 & 8 & 9 \\
Fog level & 0 & 1 & 2 & 3 & 4 & 5 & 6 & 7 & 8 & 9 \\
UAV altitude & 0.5k & 1.1k & 1.7k & 2.3k & 2.9k & 3.6k & 4.2k & 4.8k & 5.4k & 6.0k \\
\multicolumn{2}{c}{Success Threshold $\tau = 0.5$} & \multicolumn{9}{c}{\textbf{Metadata}: Detection conf. threshold $\geq$ 0.35} \\
\CodeAfter
  \tikz \draw [red, very thick] (2-|1) rectangle (7-|2);
  \tikz \draw [green!70!black, very thick] (2-|2) rectangle (3-|12);
  \tikz \draw [blue, very thick] (3-|2) rectangle (7-|12);
\end{NiceTabular}
\vspace{-7mm}
\end{table}

Each Observable (OBS) reports two values: an \textit{ethical difficulty} and a \textit{confidence value}. The ethical difficulty of the OBS is determined by simulation parameters and the OBS \textit{Key Factors} table. Table~\ref{tab:key_factors} shows a basic example of a key factor table for a UAV identifying vehicles. Each key factor, e.g., UAV's altitude or the level of rain and fog, as outlined in red, is represented by a scalar value that increases (or decreases) in proportion to its contribution to making the OBS more difficult to achieve. Each key factor has its own range and units (e.g., hours for time of day, meters for altitude, etc.). The ratings for each key factor (blue box) are mapped to a unitless, shared ethical difficulty space ranging from 0 to 100 (green box). Table~\ref{tab:key_factors} demonstrates a linear mapping, but the framework readily supports more complex models. Ethical difficulty is a property of the simulation configuration and is independent of the agent's capabilities. 
For each SimSpec, the ethical difficulty of the OBS is the \textit{minimum} ethical difficulty level across all key factors. 
This suggests that a successful trial can only certify you to the minimum of the difficulty of the trial. The end-user can tailor the relationship between simulation parameters and ethical difficulty by adjusting the key factor tables.

The confidence value of the Observable is the ratio of the number of times the agent achieved that Observable over the number of considered simulation runs. The Observables are evaluated on a binary, pass-or-fail basis using a success threshold tailored to each Observable, so the confidence value is defined over $\{0, 1\}$ for a single simulation run. The success threshold $\tau$ is defined over the continuous range $[0,1]$ and may depend on Observable-specific metadata for quantifying the behavior. From the example shown in Table~\ref{tab:key_factors}, $\tau = 0.5$ (meaning that the agent needs a positive detection on 50\% of camera frames when an object is in the field of view) while the metadata specifies that the agent must have a detection confidence of 0.35 or higher to be considered a positive detection. 

For each run, the agent receives, for each Observable, a score equal to the Observable's ethical difficulty under that simulation configuration if it meets the success criteria, and NaN otherwise.
For Observables where performance is easy to quantify across a range of values, we incorporate a partial credit score for under- or over-performance relative to the success threshold. Let $\tau_i \in [0, 1]$ be a success threshold for Observable $i$ and $x_i \in [0, 1]$ be the measured value for $i$ on a given simulation run. We assign a partial credit score, up to 4 points, using $\delta_i = 4d_{if}$, where $d_{if} = x - \tau_i$. If the difficulty for $i$ is $d_i$, then the final score for the Observable on any given run is $s_i = d_i + \delta_i$. 
For each intermediate node $v$ in the decomposition DAG, let $S = \{s_1, s_2, \cdots s_n\}$ be the scores of all child nodes and $S_s=\{ s_i \in S \mid s_i \neq \text{NaN} \}$ be the set of successful child node scores of $v$. The score for $v$ is defined as $\min ( S_s )$, and the confidence score is $\texttt{N}_{s}/|S|$, where $\texttt{N}_{s} = |S_s|$.
% Let $v$ be any node whose subgraph leaf nodes are the set of Observables $\{\texttt{Ob}_v\}$ with $\texttt{NO}_{v}=|\{\texttt{Ob}_v\}|$. Within a single run $r$, the set $S_r = \{S_{\texttt{Ob}1}, S_{\texttt{Ob}2}, \cdots S_{\texttt{Ob}j}\}$ is the set of successful Observable scores with $\texttt{NS}_{r} = |\{S_r\}|$; the ARL score for $v$ is $\min\{S_r\}$, and its confidence score is $\texttt{NS}_{r}/\texttt{NO}_{v}$. 

\begin{figure*}[!t]
  \centering
  \includegraphics[width=\linewidth]{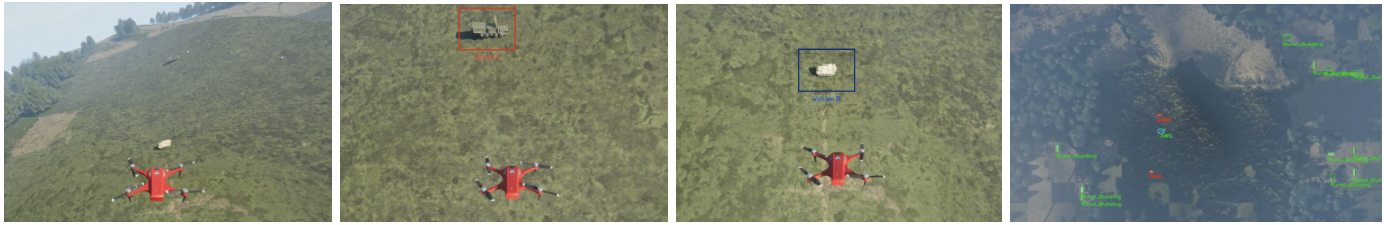}
  \caption{Example showing the UAV searching for a high-priority target (panels 1 through 3) in the FalconSim and a ``satellite'' view (panel 4) showing the layout of the environment.}
  % \caption{Priority sensitive course of action changes under controlled priority perturbations. The runs differ only in the relative priority ordering, yielding distinct search/approach decisions and resulting trajectories. }
  \label{fig:priority_paths}
  \vspace{-1em}
\end{figure*}

Table~\ref{tab:score_exp} shows a basic scoring example. In the example, Observables 1 and 2 (OBS1 and OBS2) have difficulty scores of 30 and 80, respectively, and contribute to VAB1, while OBS3 has an ethical difficulty of 50 and is the sole contributor to VAB2. Both VAB1 and VAB2 contribute up to KA1. In the first run, the agent achieved all observables, giving VAB1 a score of $min(30, 80) = 30$, VAB2 a score of 50, and KA1 a score of $min(30, 50) = 30$ with a confidence score of 1 across all attributes. The agent failed OBS1 during the second run, changing VAB1 to $min(NaN, 80) = 80$ with a confidence of only 50\%, giving the agent a higher score for VAB1 but lower confidence. The failure is further reflected up the tree, where KA1 now has a confidence score of 66.6\%. We chose this scoring approach because it independently tracks successful and failed Observables at each decomposition level, while providing statistical significance to confidence levels across repeated simulation runs.

\begin{table}[t]
\centering
\caption{Scoring example for observables, VABs, and an upstream KA}
\label{tab:score_exp}
\renewcommand{\arraystretch}{1.3} 
\setlength{\tabcolsep}{3pt} % Reduces horizontal padding slightly
% Resizebox forces the table to the exact width of a single IEEE column
\resizebox{\columnwidth}{!}{%
\scriptsize % Drops the font size so resizebox doesn't have to scale it too drastically
\begin{tabular}{|l|c|c|c|c|c|c|}
\hline
\multicolumn{7}{|c|}{\textbf{Run 1}} \\
\hline
Node & \cellcolor{red!20}\textbf{OBS1} & \cellcolor{red!20}\textbf{OBS2} & \cellcolor{green!30}\textbf{VAB1} & \cellcolor{red!20}\textbf{OBS3} & \cellcolor{green!30}\textbf{VAB2} & \cellcolor{magenta!30}\textbf{KA1} \\
\hline
Score & \textbf{30} & \textbf{80} & $\min(30, 80) \rightarrow \textbf{30}$ & \textbf{50} & $\min(50) \rightarrow \textbf{50}$ & $\min(30, 50) \rightarrow \textbf{30}$ \\
\hline
Confidence & 1 & 1 & $2/2 \rightarrow 100\%$ & 1 & 1 & $3/3 \rightarrow 100\%$ \\
\hline
\hline
\multicolumn{7}{|c|}{\textbf{Run 2}} \\
\hline
Node & \cellcolor{red!20}\textbf{OBS1} & \cellcolor{red!20}\textbf{OBS2} & \cellcolor{green!30}\textbf{VAB1} & \cellcolor{red!20}\textbf{OBS3} & \cellcolor{green!30}\textbf{VAB2} & \cellcolor{magenta!30}\textbf{KA1} \\
\hline
Score & \textit{NaN} & \textbf{80} & $\min(\mathit{NaN}, 80) \rightarrow \textbf{80}$ & \textbf{50} & $\min(50) \rightarrow \textbf{50}$ & $\min(80, 50) \rightarrow \textbf{50}$ \\
\hline
Confidence & 0 & 1 & $1/2 \rightarrow 50\%$ & 1 & 1 & $2/3 \rightarrow 66\%$ \\
\hline
\multicolumn{7}{c}{Final ARL score = 40, confidence = 83.3\% }
\end{tabular}%
}
\vspace{-7mm}
\end{table}

Let $S^N_i$ and $C^N_i$ be the set of scores and confidence, respectively, for node $i$ across a set of $N$ runs. Node $i$'s overall \textit{ARL score} and confidence are the means $\sum{S^N_i}/N$ and $\sum{C^N_i}/N$, respectively. Per Table~\ref{tab:score_exp}, the average ARL for KA1 is $\frac{30+50}{2} = 40$, while the confidence is $\frac{1.0+0.66}{2} = 0.83$.

\section{Framework Implementation Details}
\label{sec:sim_results}

% \subsection{Simulation Setup}
% \label{subsec:sim_setup}
We used the Falcon simulation environment to evaluate the autonomous agent under test. We used ROS~2 as the communication middleware between the simulator and the agent. A central design goal was to decouple the perception and reasoning modules from any particular simulator to support generalizable and reproducible evaluation across testbeds. The simulator provides standardized sensor and state streams (e.g., imagery, vehicle state). This compartmentalization enables (i) controlled substitution of perception sources (pixel-based vs.\ simulator-provided detections), (ii) consistent logging for post hoc audit, and (iii) standardized test harnesses for DoE-style evaluation.

% We evaluated the autonomous system under test in simulation using the Falcon simulation environment, with ROS~2 serving as the communication middleware between the simulator and the system modules. A central design goal was to decouple the perception and reasoning modules from any particular simulator to support generalizable and reproducible evaluation across testbeds. Concretely, the simulator provides standardized sensor and state streams (e.g., imagery, vehicle state), while the system returns a compact set of outputs (e.g., navigation and action decisions) through a fixed input--output interface (Figure~\ref{fig:aws_dataflow}). This compartmentalization enables (i) controlled substitution of perception sources (pixel based vs.\ simulator provided detections), (ii) consistent logging for post hoc audit, and (iii) standardized test harnesses for DOE style evaluation.

% \begin{figure}[!ht]
%   \centering
%   \includegraphics[width=\linewidth]{14.jpeg}
%   \caption{Data flow between the simulation environment and the system under test. Perception and reasoning are decoupled from the simulator, enabling portability across simulation backends that adhere to a standard input--output interface.}
%   \label{fig:aws_dataflow}
% \end{figure}

% \subsection{Scenario generation for ethically relevant variability}
% \label{subsec:scenario_gen}
We leveraged a Generative Environment (GE) pipeline capable of producing variable scenarios on demand. The GE composes scenarios from (a) a curated set of plugin digital twins provided by the simulator and (b) constraint-guided scene generation. Scenarios are specified using Section~\ref{sec:3b} and created using Section~\ref{sec:3c}. This workflow supports controlled variation (including systematic perturbations) while preserving traceability and repeatability across runs, and it is designed to integrate with larger DoE pipelines (including parallel execution on compute clusters).

% Importantly for robot ethics evaluation, the GE embeds ethical considerations at the asset and constraint levels, enabling scenario families that stress test issues such as contextual sensitivity, discrimination risks (e.g., presence of bystanders), and operational uncertainty (e.g., unavailability of resources), while maintaining semantic coherence and experimental control.

% \subsection{Logging, traceability, and post hoc analysis}
% \label{subsec:logging}
Across all runs, the system produced structured logs to support transparency and post-mission audit. Logged data include raw odometry (e.g., state derived from IMU data), structured perception outputs for recognized objects, and decision outputs produced by the reasoning engine. This instrumentation is designed to enable: (i) reconstruction of when and why decisions were taken, (ii) debugging and failure analysis (e.g., distinguishing perception vs.\ reasoning failures), and (iii) downstream scoring and evaluation pipelines that compare ground truth and perception-based outputs.

% \subsection{Limitations and ethics relevant implications }
% \label{subsec:limitations}
% The present results are primarily functional and qualitative, demonstrating modular portability across perception sources, robustness demonstrations under representative environmental variation, and interpretable priority driven behavior changes. A quantitative statistical evaluation (e.g., calibrated error rates under controlled perturbations, robustness curves over scenario families, and disaggregated performance across ethically relevant subpopulations) is a natural next step enabled by the GE+DOE workflow, but is not reported in this phase. Nonetheless, the architecture and logging choices directly support core responsible robotics goals: repeatable testing, controlled uncertainty injection, and traceable decision records suitable for technical and ethical scrutiny.

% =========================
\section{Framework Evaluation}
\label{sec:experiments}
To demonstrate the utility of our REBAR framework for scoring an autonomous system's ethical behavior, we implemented a simple target-search-and-report algorithm for a quadrotor UAV in a defense application. 
The quadrotor must scan an area of interest to locate and report on assets (\eg, a mobile radar system). 
For the ethical decomposition, we used our five ethical ARLs presented in Section~\ref{sec:ARLs} for AI Principles. We further factored these five Principles into 43 KAs, 121 VABs, and 135 unique Observables. The nodes in the DAG have a many-to-many relationship. For example, a single KA could be composed of multiple VABs (shown in Figure~\ref{fig:rebar_graph}) and contribute to multiple Principles.
% and that appeared 158 times as children of multiple VABs. 
The rest of the section covers the details of how we implemented the autonomous agent, the experiment sets used, and the results of these experiments. 

% We evaluated the quadrotor's autonomy stack in FalconSim using the interface and instrumentation described in Sec.~\ref{subsec:sim_setup} (Fig.~\ref{fig:aws_dataflow}), with perception and reasoning components as in Secs.~\ref{subsec:perception_eval}--\ref{subsec:reasoning_eval}.

% Following a batch of simulation runs, the framework conducts an automated statistical analysis on the key factors to determine their actual impact on the resulting ARL scores. This analysis includes:
% \begin{itemize}
%     \item ANOVA (Analysis of Variance): To identify which factor levels significantly alter an autonomous system's ethical performance.
%     \item Linearity and Independence Testing: Utilizing Pearson correlation to assess factor independence.
%     \item Jensen-Shannon (JS) Divergence: Used to quantify the distributional differences in ARL scores between experimental iterations, precisely measuring how system behavior shifts when ethical parameters are modified.
% \end{itemize}

% By employing a human-in-the-loop review of these statistical artifacts, researchers can iteratively drop insignificant variables, adjust parameter bounds, and generate subsequent experimental designs that hone in on the precise boundaries of an autonomous system's ethical competence.

\begin{figure}[!b]
  \centering
  \includegraphics[width=\linewidth]{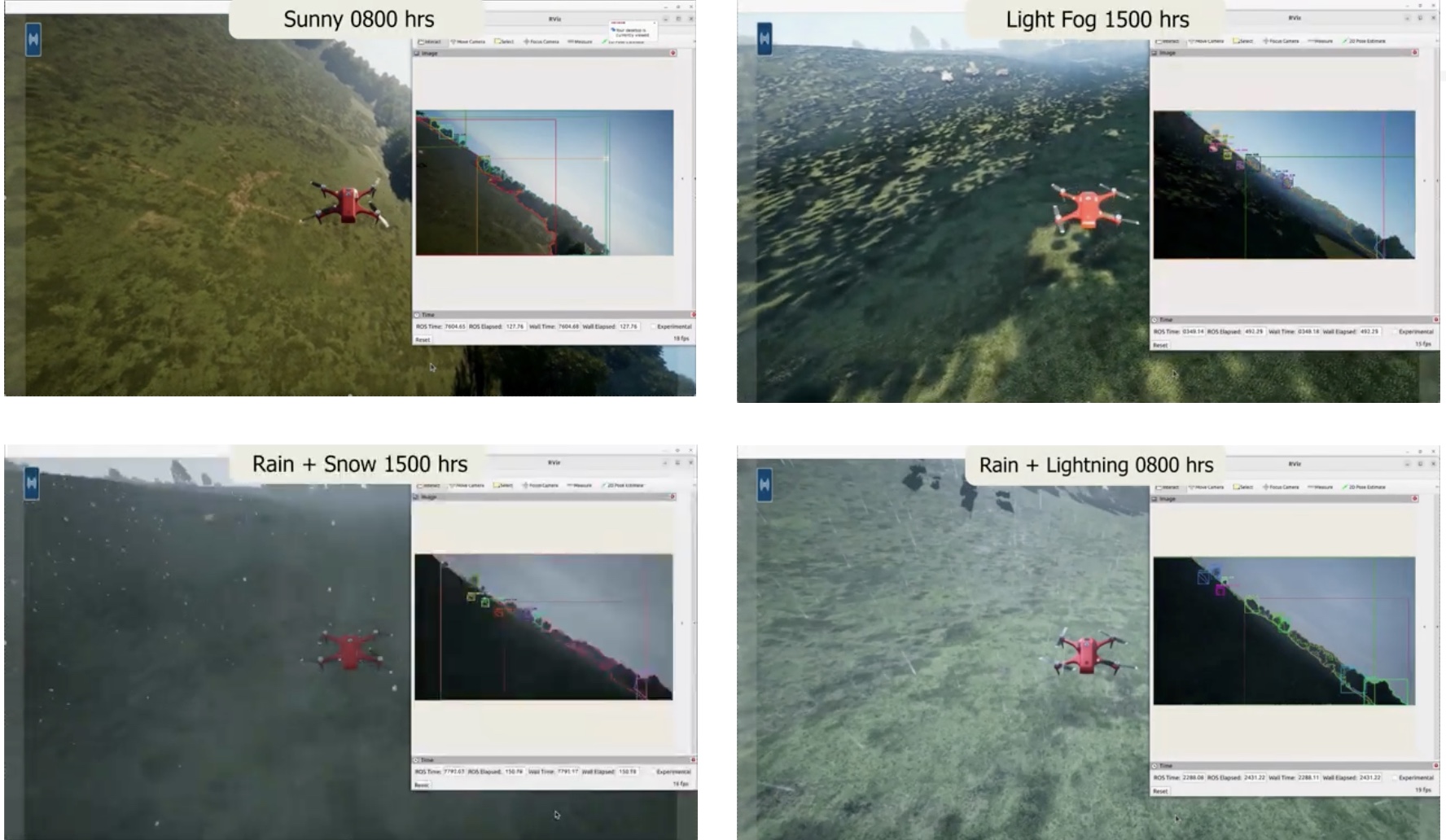}
  \caption{Representative examples of the pixel-based perception pipeline identifying relevant objects under varying illumination and weather conditions, and supporting downstream autonomous decisions in simulation.}
  \label{fig:perception_examples}
\end{figure}

\subsection{Autonomous Agent Implementation}
% \label{subsec:reasoning_eval}
Building on prior work \cite{Ravichandran2025}, the reasoning engine was evaluated for intelligent search and prioritization under incomplete and time-sensitive cues. The system receives approximate location intelligence, target priorities, and cue latency information, and selects which objectives to pursue first while accounting for limited resource availability. In simulation, mission execution followed a consistent three-stage structure: (1) move to the subregion of the environment based on a target location intelligence report, (2) search that subregion using a lawnmower search pattern, and (3) identify/engage targets when specific conditions are met. Engagement conditions include object detection confidence, multiple observations from varying viewpoints, and consistent object labeling across consecutive frames.

A key qualitative result is that the system exhibits priority-sensitive course of action changes under controlled priority perturbations. Figure~\ref{fig:priority_paths} illustrates two runs in which only the relative priority ordering differs, yielding distinct search/approach decisions and resulting trajectories. This behavior is particularly relevant to ethics and governance discussions because it reveals how explicit mission priorities and resource constraints translate into observable system behavior, which can then be interrogated for compliance, proportionality, and operator expectations.

\begin{figure*}[!ht]
  \centering
  % Can we get the original graphs?
  \includegraphics[width=\linewidth]{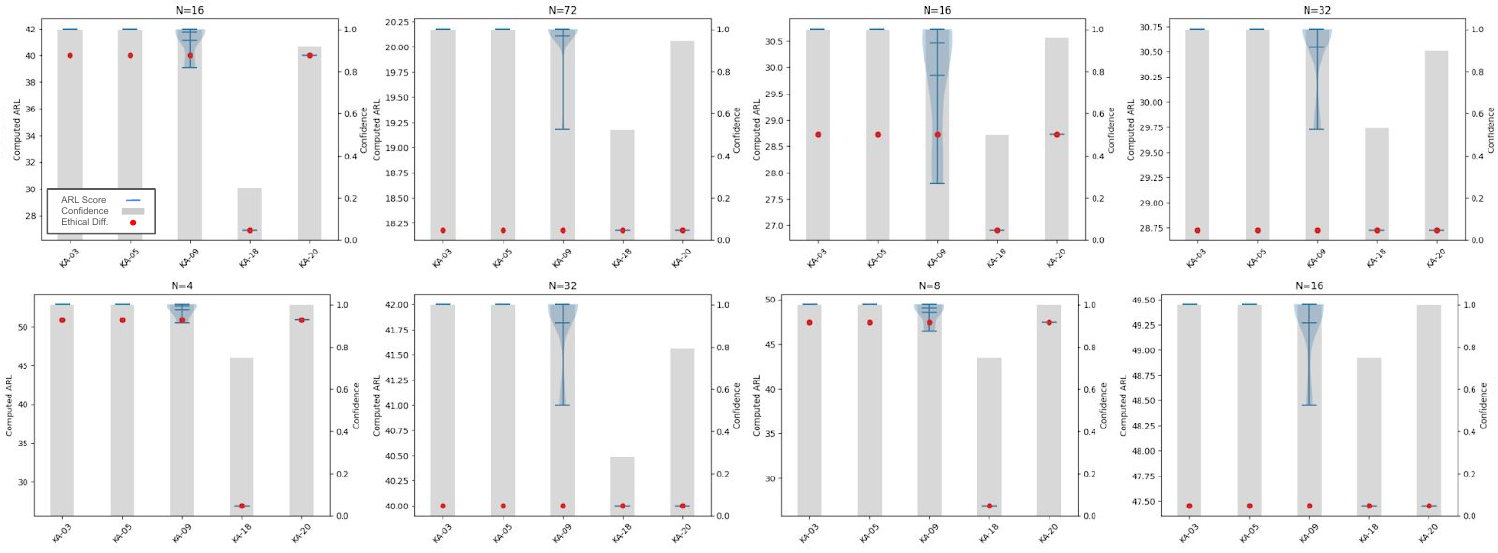}
  \caption{Final ARL scores after $N$ runs for KA-03 (bystander classification), KA-05 (adversary classification), KA-09 (object detection), KA-18 (bystander proximity reasoning), and KA-20 (mission accomplishment), showing a high task success rate but a failure to reason about an action's impact on bystanders.}
  \label{fig:arl_results}
  \vspace{-1em}
\end{figure*}

% \subsection{Results}
% \label{sec:results}

% \subsection{Perception evaluation in simulation}
% \label{subsec:perception_eval}
% The perception module was evaluated in two complementary modes. First, in a pixel based configuration, 
% To create a vision pipeline for the agent, 
% we used GroundedSAM (Grounding DINO + SAM) \cite{Tianhe2024} to support open set recognition and segmentation via textual prompts. This enables prompt driven category specification without task specific retraining, and supports rapid updates when new or partial instructions become available. 
Figure~\ref{fig:perception_examples} shows examples of the vision pipeline drawing bounding boxes around identified objects in varying weather conditions.
% In representative simulated conditions, the pixel based pipeline successfully identified relevant objects under varying illumination and weather conditions consistent with the ``Tension~1'' stressors described in the program context (the tradeoff between completing missions and minimizing civilian interaction (e.g., increased ambiguity from adverse weather and target--civilian similarity)---as described in the program context (examples in Figure~\ref{fig:perception_examples}).
The module distinguishes between intended objects of interest and civilian bystanders, which is critical for subsequent ethical analysis of misidentification and downstream decision-making.

\begin{table}[b]
\centering
\caption{Environment Configuration Parameters}
\label{tab:env_params_horiz}
\setlength{\tabcolsep}{3pt} % Reduces padding between columns
\resizebox{\columnwidth}{!}{%
\begin{tabular}{|c|c|c|c|}
\hline
\textbf{Approach} & \textbf{Concealment} & \textbf{Civ. Density} & \textbf{Civ. Proximity} \\ \hline
 \{N, E, S, W\} & \{0.5, 1\} & \{313.3, 390.1, 453.3\} & \{229.9, 259.1, 362.4\} \\ \hline
\end{tabular}%
}
\end{table}

% Second, to isolate and validate downstream reasoning behaviors independent of pixel level failures, we evaluated the system using simulator derived detections ijected with noise to mimic real detection uncertainty. This mode provides a controlled pathway to test prioritization, search, and decision logic while systematically varying detection noise characteristics.

\subsection{Experiment sets}

To test the capabilities of the autonomous agent for military applications, where the agent must reason about the existence of innocent bystanders, we generated simulation configurations using the environment ranges listed in Table~\ref{tab:env_params_horiz}. We generated 215 different SimSpecs by sampling unique combinations from the listed ranges, which yielded 12 unique ethical difficulty levels. 

To demonstrate how multiple simulation configurations can give us a subset of unique ethical difficulties, we refer back to the example in Table~\ref{tab:key_factors}. A SimSpec with \textit{Hours till solar noon} = 5, \textit{Rain level} = 1, \textit{Fog level} = 2, and \textit{UAV altitude} = 3k would have a difficulty of 20 because a \textit{Rain level} of 1 is the lowest rating in the Key Factor table of all settings and corresponds to a difficulty rating of 20. If we create another SimSpec that has \textit{Rain level} = 2 and \textit{Fog level} = 1, with the other settings remaining the same, the difficulty would again be 20, but is now dictated by the \textit{Fog level} setting instead of the \textit{Rain level} setting.

We focus on five KAs from our decomposition to demonstrate how the REBAR framework assessed the autonomy agent's performance. These KAs are: classification of civilian bystanders and vehicles (KA-03), classification of enemy personnel and vehicles (KA-05), object detection (KA-09), suppression of target-marking near bystanders (KA-18), and mission accomplishment (KA-20).

Figure~\ref{fig:arl_results} shows scores for eight of the twelve unique ethical difficulty configurations.
% ; i.e.\ each subplot's configuration of red dots (ethical difficulties) is unique, but may arise from different simulation configurations. 
The instance count for each configuration is shown as $N$. The red dot and blue violin plot show the ethical difficulty and ARL score, respectively, for each KA, measured along the left-hand \textit{y-axis}.
% Each KA score is represented by a red dot indicating the ethical difficulty and a blue violin plot showing the range of acheived ARL scores (left Y-axis) and a 
The gray bar indicates the confidence of the ARL scores, measured along the right-hand \textit{y-axis}. 
The ARL scores vary from the ethical difficulty due to a partial credit mechanism. 
% which awards an bonus for Observables that exceed their success criteria, or partial ARL credit for failed Observables which came close to succeeding.

These results show that the UAV performs well when classifying mission-relevant objects (KA-03 and 05) and general environmental objects (KA-09) over varying difficulty levels, leading to a high mission accomplishment rate (KA-20). However, the confidence score for not marking targets near civilian bystanders (KA-18) is low, even in configurations with low ethical difficulty, indicating that the agent should not be deployed in areas likely to have civilian bystanders.

\section{Conclusions \& Limitations}

% Overall, REBAR provides a transparent, repeatable path from abstract principles to independently verifiable measurements, and establishes a foundation for further work in expanding observable coverage, strengthening reasoning and intent assessment, and scaling evaluation across domains, mission types, and multi agent settings. \section{CONCLUSIONS}
\label{sec:conclusions}

% We presented REBAR, a test and evaluation framework that turns ethical and legal considerations into measurable, repeatable evidence for autonomy. REBAR moves beyond qualitative checklists by (i) exercising systems with controlled scenario families that surface ethically relevant failures, and (ii) producing independently verifiable scores from instrumented runs, including repeatability based confidence. REBAR makes evaluation concrete: it specifies \emph{what} to measure, \emph{why} it matters (Principles), and \emph{how} it is computed (success criteria and aggregation). By grounding assessments in observable behavior and traceable logs, it reduces subjectivity relative to narrative or nonauditable evaluations and supports ethics-informed risk analysis for robot deployments. Additionally, REBAR is designed to be portable. Scenario specifications, simulator agnostic logging requirements, and a modular scoring interface enable integration with other autonomy stacks and simulation environments, providing a reusable benchmarking layer that can extend across domains, mission types, and evolving legal requirements.

We presented REBAR, an evaluation framework that translates ethical and legal principles into measurable, repeatable scores for autonomous systems, grounded in their observable behavior. 
% By replacing qualitative checklists with controlled scenarios and instrumented runs, REBAR generates independently verifiable scores grounded in observable behavior. 
% This approach significantly reduces subjectivity and supports robust, ethics-informed risk analysis. 
The framework’s modular design provides a portable benchmarking layer that seamlessly integrates with various autonomy stacks across diverse domains and evolving legal requirements. We demonstrated REBAR’s utility in identifying an autonomous agent's limitations within a defense application, highlighting its failure to account for bystander safety despite high task proficiency. Building on this initial sampling, future work will evaluate the framework across new scenarios and a variety of agents.

% Although REBAR seeks to fill a much needed gap in robot ethics research, it still contains several limitations. Constructing the ethical decomposition is labor intensive and requires both application domain knowledge and expertise in embodied AI. Furthermore, updates to the decomposition require updating the test ports that feed the framework's scoring logic by parsing simulation logs -- another labor intensive task. One possible solution for this limitation is to leverage generative AI for ethical decompositions, but grounding generative models is an ongoing topic of research. There are also factors that feed into ethical principles that are not quantifiable in a simulation. For example, an robot's resistance to physical, adversarial manipulation or the reliability of a ``kill switch'' on physical hardware.

While REBAR addresses a critical need in the evaluation of robot ethics, it has notable limitations. Developing and updating the ethical decompositions, along with the corresponding simulation log parsers, is highly labor-intensive and demands dual expertise in embodied AI and the target domain. Although generative AI offers a potential pathway to automate this, grounding these models remains an ongoing challenge. Additionally, REBAR is constrained by factors that cannot be quantified in simulation, such as hardware kill-switch reliability and physical resilience to adversarial manipulation.

% Future work could focus on (i) expanding and validating the set of online Observables implemented in simulation (e.g., broader classification targets beyond vehicles) while tightening traceability from scenario parameters to Key Factors and success criteria; (ii) improving evaluation of internal decision processes via Reasoning Observables, including causal traceability from sensor inputs to course of action selection and abort/replan decisions; (iii) extending scenario generation and SimSpec based realization creation to richer environments and additional domains (e.g., ground-to-ground and air-to-air) through semi-automated ``expansion pack'' style adaptations with SME validation; and (iv) scaling experimentation through DOE-guided sampling and parallel execution to better characterize performance--risk trade-offs across ethical difficulty levels and operational conditions.

% \addtolength{\textheight}{-12cm}   % This command serves to balance the column lengths
                                  % on the last page of the document manually. It shortens
                                  % the textheight of the last page by a suitable amount.
                                  % This command does not take effect until the next page
                                  % so it should come on the page before the last. Make
                                  % sure that you do not shorten the textheight too much.

%%%%%%%%%%%%%%%%%%%%%%%%%%%%%%%%%%%%%%%%%%%%%%%%%%%%%%%%%%%%%%%%%%%%%%%%%%%%%%%%

\section*{ACKNOWLEDGMENT}

This research was developed with funding from the Defense Advanced Research Projects Agency (DARPA). The views, opinions and/or findings expressed are those of the author and should not be interpreted as representing the official views or policies of DARPA or the U.S. Government.

%%%%%%%%%%%%%%%%%%%%%%%%%%%%%%%%%%%%%%%%%%%%%%%%%%%%%%%%%%%%%%%%%%%%%%%%%%%%%%%%

\bibliographystyle{IEEEtran}
\bibliography{references}

\end{document}